\documentclass[10pt,twocolumn,letterpaper]{article}

\usepackage{wacv}
\usepackage{times}
\usepackage{epsfig}
\usepackage{graphicx}
\usepackage{amsmath}
\usepackage{amssymb}

\def\bmvaOneDot{.\ }
\def\ie{\emph{i.e}\bmvaOneDot}
\def\eg{\emph{e.g}\bmvaOneDot}

\def\etal{\emph{et al}\bmvaOneDot}

\def\wacvPaperID{****} 

\wacvfinalcopy 

\ifwacvfinal
\def\assignedStartPage{1} 
\fi

\ifwacvfinal
\usepackage[breaklinks=true,bookmarks=false]{hyperref}
\else
\usepackage[pagebackref=true,breaklinks=true,colorlinks,bookmarks=false]{hyperref}
\fi

\ifwacvfinal
\else
\fi

\begin{document}

\title{MS-RANAS: Multi-Scale Resource-Aware Neural Architecture Search}

\author{Cristian Cioflan\\
ETH Zurich\\
{\tt\small cioflanc@student.ethz.ch}
\and
Radu Timofte\\
ETH Zurich\\
{\tt\small timofter@vision.ee.ethz.ch}
}

\maketitle

\begin{abstract}
Neural Architecture Search (NAS) has proved effective in offering outperforming alternatives to handcrafted neural networks. In this paper we analyse the benefits of NAS for image classification tasks under strict computational constraints. Our aim is to automate the design of highly efficient deep neural networks, capable of offering fast and accurate predictions and that could be deployed on a low-memory, low-power system-on-chip. The task thus becomes a three-party trade-off between accuracy, computational complexity, and memory requirements. To address this concern, we propose Multi-Scale Resource-Aware Neural Architecture Search (MS-RANAS). We employ a one-shot architecture search approach in order to obtain a reduced search cost and we focus on an anytime prediction setting. Through the usage of multiple-scaled features and early classifiers, we achieved state-of-the-art results in terms of accuracy-speed trade-off.
\end{abstract}

\section{Introduction}

Tens of thousands of images are captured every second by the world's population, which led to an increased interest in image classification tasks. Subsequently, remarkable progress has been made in designing systems that can achieve satisfactory results on object recognition benchmark datasets, such as ImageNet~\cite{imagenet} or COCO~\cite{coco}. Nowadays, there exists a paradigm shift regarding where the inference process takes place. Specifically, instead of using multi-node clusters for simple object recognition tasks, more applications~\cite{edge,ai} are designed following the edge computing principle, thus bringing data storage and computation closer together, hence saving bandwidth and improving response time.

In an edge computing setting, three main metrics should be taken into account when evaluating a classification system, namely the accuracy in solving a specific task, the computational latency, and the on-chip memory size of the deployed model. Only by considering these three metrics one could create a model able to offer fast and accurate predictions under low-memory, low-power constraints.

The design of efficient, accurate neural networks is a challenging task for computer vision experts in this day and age. Choosing the adequate network architecture has proven to be crucial for the system's performance and, when one has to take into account multiple metrics, the process becomes all the more laborious. Because of this, a rapidly growing research topic addresses the development of algorithms capable to automate the discovery of optimal architectures, namely neural architecture search (NAS) algorithms~\cite{survey}.

Our main focus is represented by the trade-off between accuracy and latency. We therefore address the problem of \textit{anytime prediction}, the network's ability to offer predictions at any given point. We propose Multi-Scale Resource-Aware Neural Architecture Search (MS-RANAS), a novel approach towards obtaining efficient neural networks, capable of successfully accomplishing the aforementioned tasks. The tedious challenge of handcrafting the structure of the network is alleviated through the usage of NAS algorithms, while the quality and complexity of the features passed throughout the network are preserved by introducing a multi-scale structure. Lastly, the latency issue is managed through the usage of early exits.

\begin{figure*}[th!]
\begin{center}

\includegraphics[width=0.8\linewidth]{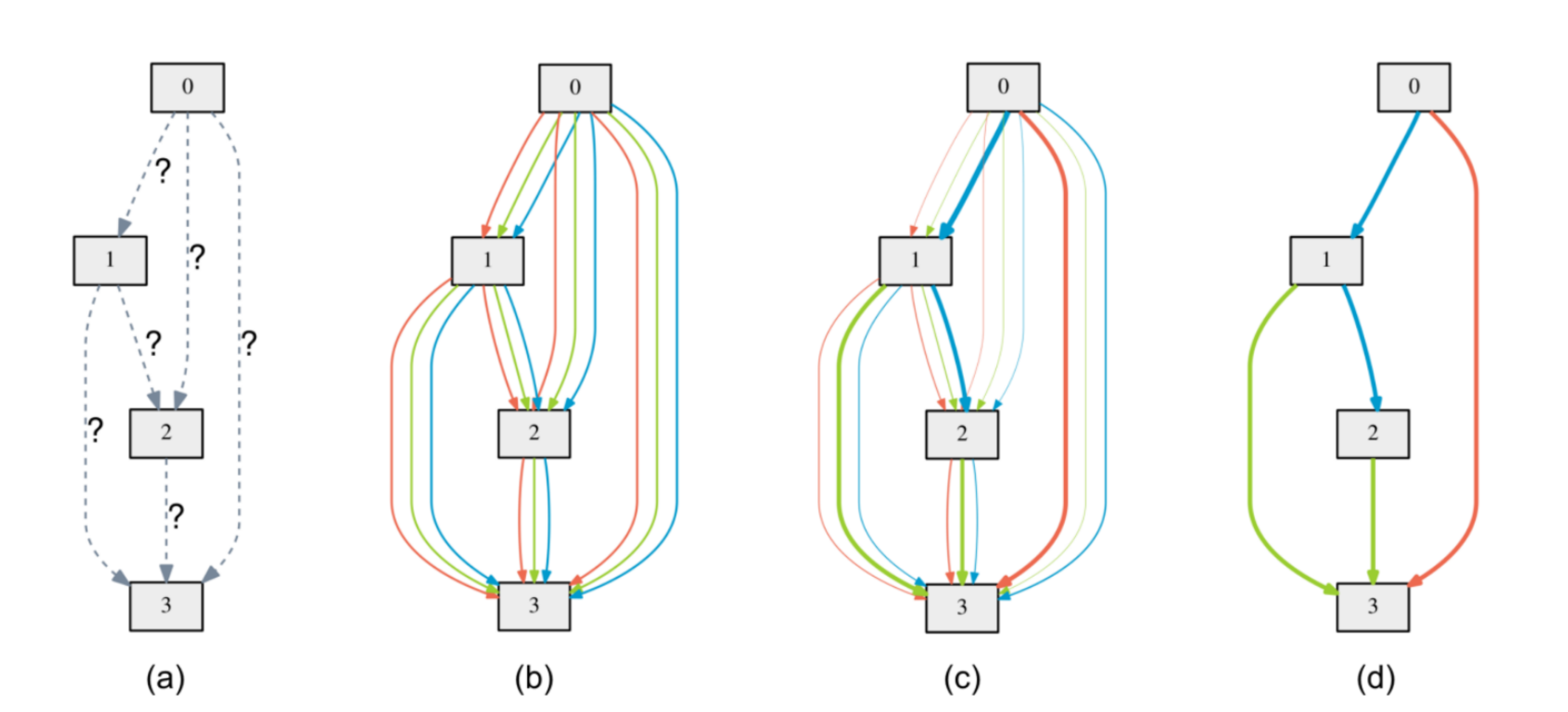}
\end{center}
      \caption{\textbf{DARTS~\cite{darts} overview:} (a) Operations between nodes are unknown in the first step; (b) Continuous relaxation allows the placement of a candidate operations mixture on each edge; (c) Joint optimisation of the mixing probabilities and of the network weights by finding the solution of a bilevel optimisation problem; (d) Inferring the final architecture using learned mixing probabilities. Figure from~\cite{darts}.
     }
\label{fig:darts}
\end{figure*}
\section{Related work}

\subsection{Neural architecture search}

Several approaches have been made in order to automate the generation of robust neural networks. These approaches include the use of reinforcement learning~\cite{zoph,baker}, evolutionary algorithms~\cite{real}, and surrogate model-based optimisations~\cite{bayesian,encoder}. Nevertheless, the prolonged searching time~\cite{survey} might be unfeasible in particular scenarios, we are thus compelled to use a different method.

One-shot architecture search is defined by Witsuba~\etal~\cite{survey} as a system which, during the entire search process, trains a single neural network, used afterwards to derive different architectures throughout the search space as possible candidates to the optimisation problem that was imposed. The neural architecture search space represents a subspace that includes all the nodes and the set of operations that have to be applied by each node in order to compute its output. As certain constraints can be imposed on the final architecture, the concept of search space can be restricted to the set of feasible solutions such a NAS technique may return. The search space can be divided into two classes: a global search space, where a graph is used to define the entire architecture, and a cell-based search space, in which a several cells are replicated through the architecture to form the final network. Empirical evidence provided by Pham~\etal~\cite{pham} shows that the cell-based approach outperforms the global one both in terms of accuracy and in number of parameters.

Subsequently, Liu~\etal~\cite{darts} proposed differentiable architecture search (DARTS, see Figure~\ref{fig:darts}). As opposed to previously researched techniques, which involved applying reinforcement learning or using evolutionary algorithms over a discrete, non-differentiable search space, DARTS introduces a continuous relaxation of the architecture representation, thus favouring the use of gradient descent for a more efficient search of the network architecture. The relatively short GPU time necessary to discover an efficient architecture (\ie 4 GPU days) and its high accuracy, together with the comparably small amount of parameters and the modularity of the structure per se, which results from the cell-based search space approach, are the main reasons for choosing this method as the foundation of our work.

\subsection{Latency reduction}
\label{ssc:latency}

Recent hardware and algorithmic advances have stimulated the design of networks with increased depths, with the end goal of obtaining a better performance. The result was an explosion in number of layers, and in number of parameters implicitly, from VGG-16~\cite{vgg}, which achieved 74.4\% accuracy on ImageNet dataset~\cite{imagenet} with 138 million parameters, to ResNeXt-101 32x48d~\cite{resnext}, with 85.4\% accuracy using 829 million parameters. Apart from the on-chip memory that these fast growing networks take up, one has to take into account other factors as well, such as the latency and the energy required for feedforward inference. In many real world scenarios, and especially in an edge computing setting, the end-user is willing to trade several percentages in accuracy in order to consume less energy and to have a shorter runtime. 

Different methods have been used to alleviate the latency issue and, inherently, to reduce the power consumption. Whilst initially introduced to alleviate the vanishing gradient problem, the method of integrating skip connections (\ie dynamically adjusting which layers of a network are to be executed during inference) was successfully used to reduce the inference time as well.~\cite{act,sact,skipnet}. While these models successfully address the latency issue, they generate an overhead in terms of network parameters, therefore a different approach has to be taken in order to obtain a satisfactory trade-off between accuracy, latency, and memory.

\begin{figure*}[th!]
\begin{center}

\includegraphics[width=1\linewidth]{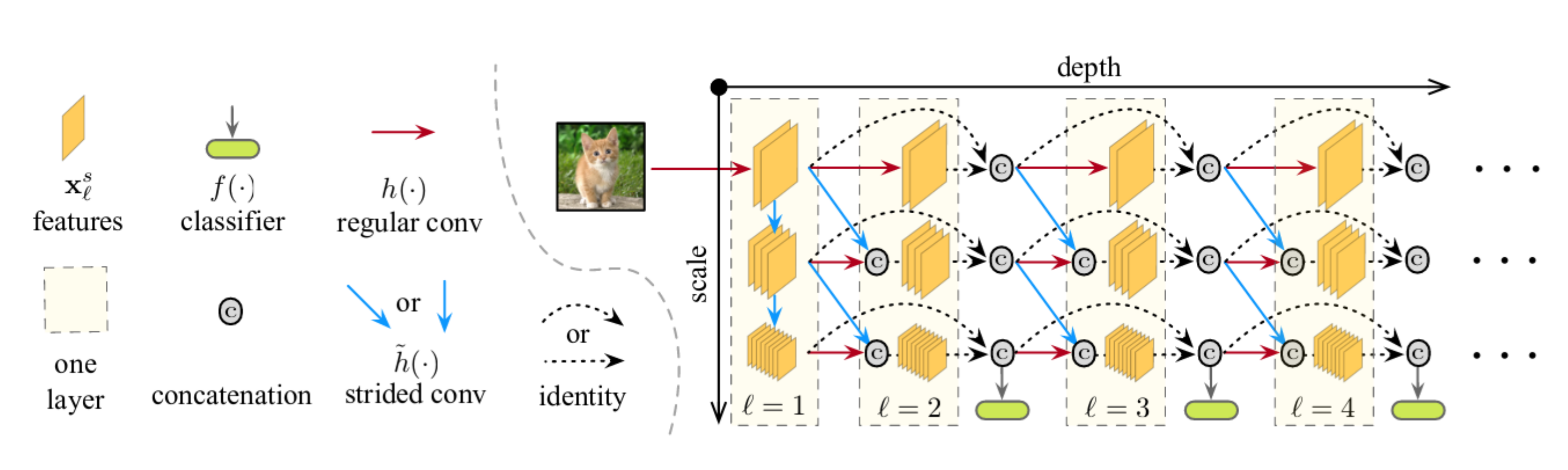}
\end{center}
      \caption{\textbf{MSDNet~\cite{msdnet} overview:} the depth of the network (\ie layers) is represented in the horizontal direction and the scales of the feature maps are represented vertically. Classifiers use the coarsest scale to compute their predictions. Figure from~\cite{msdnet}.
   }
\label{fig:MSDNet}
\end{figure*}

The second approach aimed at reducing the latency in a deep neural network is represented by early exits. This assumes the placement of additional classifiers after intermediate layers of the network, thus allowing the system to finish the execution before reaching the final layer. Subsequently, the final classifier is expected to be reached only for complex data, which results in a reduced computational effort. Huang~\etal~\cite{msdnet} presented Multi-Scale Dense Networks (MSDNet), shown in Figure~\ref{fig:MSDNet}, where they follow the principle of anytime inference. Their work addresses a relevant issue of early exits, namely the fact that the first layers of a deep neural network extract fine-grained features, whilst the deep ones operate with coarse-grained features, thus the scale of the features throughout the network is varying. In order to alleviate this problem, the authors propose maintaining features of all scales at any point in the network. By always having both fine- and coarse-grained features to learn from, classifiers can use the coarse-level ones to operate their predictions, whilst the others can be used as the main source of knowledge propagation. Early exits are used as well in the work of Zhang~\etal~\cite{ghn}, who proposed Graph HyperNetwork (GHN). Their approach relies on neural architecture search, as opposed to the handcrafted network presented by Huang~\etal~\cite{msdnet}. A mixture of graph neural networks and hypernetworks is used, thus allowing for the generation of weights based on the computation graph representation of the architecture. 

\subsection{Model compression}
\label{ssc:model_compression}

As discussed in subsection~\ref{ssc:latency}, the goal of designing high-accuracy networks led to deeper and deeper architectures. The clear trade-off of this scenario consists in the uprising of memory expensive and computationally intensive models. Nevertheless, several methods could be performed in order to help compressing the models, while simultaneously preserving their performance. According to Cheng~\etal~\cite{compression}, these techniques can be divided in four classes: (1) parameter pruning and sharing based methods, which make use of the model parameters' redundancy, trying to remove the redundant and unimportant ones, (2) low-rank factorisation based methods, that use tensor decomposition to estimate the valuable parameters of the network, (3) the transfer or the compression of convolutional filters, aimed at reducing the parameter space, and (4) knowledge distillation techniques, allowing for larger model to be reproduced using a shallower one. Additionally, we have to mention quantisation~\cite{inq}, the reduction in number of bits of a network's parameters, thus reducing both bandwidth and storage requirements with little to no drop in performance.

\section{Multi-Scale Resource-Aware Neural Architecture Search}
\label{sec:MS-RANAS}

\subsection{Differentiable architecture search}
\label{ssc:differentiable}
Following Liu~\etal~\cite{darts}, we propose the search of a computation cell as a building block that is replicated throughout the final architecture. From that perspective, a cell is a directed acyclic graph comprising an ordered set of $N$ nodes, a description that follows Bauer's formalisation~\cite{bauer}.

\begin{figure*}[th!]
\begin{center}

\includegraphics[width=0.95\linewidth]{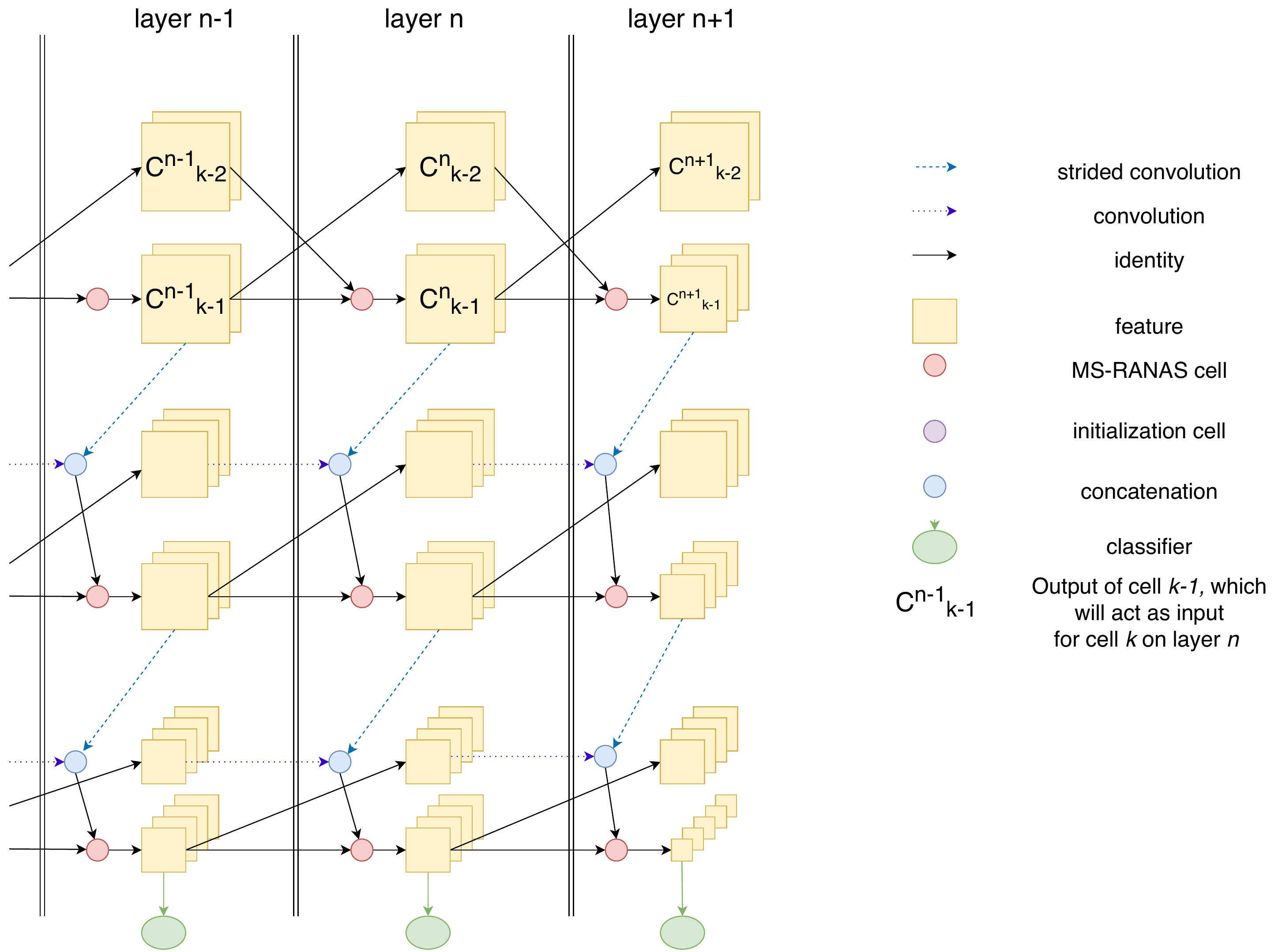}
\end{center}
   \caption{\textbf{MS-RANAS (ours) overview:} The connections between any two layers. The first two layers are common ones, whilst layer $n+1$ is a reduction layer, which contributes to decreasing the spatial dimensionality.}
\label{fig:msranas}
\end{figure*}

Each node $x^{(i)}$ represents a feature map and is associated with an operation $o^{(i,j)}$ that acts as a directed edge $(i, j)$ between the node that will be processed (\ie $x^{(i)}$) and the node's child (\ie $x^{(j)}$). Instead of assuming that each operation from the search space $K$ is either part of the final network or it is not (\ie $\alpha_{i,j}\in\{0,1\}$), Liu~\etal~\cite{darts} propose a relaxation of this assumption, namely they consider a linearly weighted combination in which $\alpha_{i,j}$ can take any real value in the range $[0,1]$, which admits for a soft decision on each path. Considering $K$ as the set of candidate operations, where each operation is a function $o(\cdot)$ that will be applied on $x^{(i)}$, the relaxation of the categorical choice of a specific operation to a softmax over all possible operations leads to the following equation:

\begin{equation}
\label{eq:soft}
\bar{o}^\mathrm{(i,j)}(x) = \sum_{k \in K} \dfrac{exp(\alpha^{(i,j)}_k)} {\sum_{k' \in K} exp (\alpha^{(i,j)}_{k'})} o(x)
\end{equation}

The softmax performed in Equation \eqref{eq:soft} ensures that for every $k$
\begin{equation}
\label{eq:sum}
\sum_{i<j} \sum_{k \in K} \alpha^{(i,j)}_k = 1 .
\end{equation}

Therefore, the task of searching for a specific architecture can be reduced to learning a set of continuous variables $\alpha = \{\alpha^{(i,j)}\}$, as shown in Figure~\ref{fig:darts}. Thus, we can consider $\alpha$ to be the encoding of the architecture, or the architecture itself.

\subsection{Multi-scale feature maps}
\label{ssc:multiscale}
As we presented in subsection~\ref{ssc:latency}, the memory-favourable approach that can be used in order to reduce the latency of a network is represented by the introduction of early exits, thus having accurate-enough predictions with a short runtime and going through the entire set of layers only for complex inputs. Given the results obtained by Huang~\etal~\cite{msdnet} through the usage of multi-scaled feature maps, we will implement a similar technique in the current work. Namely, we will have up to three scales, the larger one having the same size as the input, whilst the height and the width of the rest will be halved in comparison to predecessor. The convolution of features of similar size from one layer to another will be herein after named \textit{same-scale convolution} or \textit{horizontal convolution}. Secondly, in order to go from a fine-level feature to a coarser one, we will apply a strided convolution, referred to as \textit{diagonal convolution}. Additionally, we will use the term \textit{horizontal level}, or simply \textit{level}, representing all the equally-sized feature maps through the entire network. For the upper level, where the features have the same size as the input, we will use only same-scale convolutions. For all the other levels, the input of an MS-RANAS cell (\ie a layer derived using our neural architecture search technique) will be represented by a concatenation between a diagonal convolution of the previous larger-scale feature and a horizontal convolution of the previous same-scale feature. In addition, the first layer will be used only to generate all the multiple scales, through a strided convolution, known as a \textit{vertical convolution}. To explain concisely, horizontal convolutions help preserving and feeding forward high-resolution information, which allows for the creation of relevant coarse-grained features in deep layers, whilst the diagonal convolutions facilitate the access of classifiers to both fine- and coarse-level features. An overview of MS-RANAS can be observed in Figure~\ref{fig:msranas}.

\subsection{Learning procedure}
\label{ssc:learning}
The final network is comprised of stacked identical cells, each of them learning its own parameters. Unless stated otherwise, we are using four intermediate nodes, each with two inputs, whose output is a simple addition of the inputs. The sole purpose of these nodes is to create the context for intermediate operations, thus allowing a variety of possible interactions of the inputs from one layer to another. The complete set of operations that can take place between the inputs is known in literature as the NASNet search space~\cite{nasnet}. The reason behind using depth-wise separable convolutions (\ie a depth-wise convolution, followed by a point-wise convolution) and dilated convolutions (\ie a convolution with a larger receptive field) lies in the lower amount of multiplications and additions than the common 2D convolution. The final output is obtained as a concatenation of all intermediate nodes' results.

During both searching for an architecture and its training, we are using cross-entropy loss function $L(c_{k})$ for all the classifiers. Our goal is to minimise the weighted cumulative loss:
\begin{equation}
\label{eq:loss}
L = \frac{1}{\|D\|} \sum_{(x,y) \in D} \sum_{k} w_{k}. L(c_{k})
\end{equation}
Here, $D$ refers to the training set and $w_{k}$ is the weight of the $k$-th classifier. Intuitively, having a larger weight for the early classifiers would lead to more precise result in the first layers. Yet, given the conclusions of Huang~\etal~\cite{msdnet} and the fact that we want to limit the constraints imposed on the searching algorithm, we decided to have an equal weight for all the predictors.

\section{Experiments and results}
\label{sec:experiments}

Unless stated otherwise, our experiments~\footnote{All our codes and models will be released.} were conducted on CIFAR-10~\cite{cifar10}. Both the architecture search phase and the training one are following the methodology proposed by Liu~\etal~\cite{darts}. 

\subsection{Enhancing the search process}
\label{ssc:search_process}

\begin{figure}[th!]
\begin{center}

\includegraphics[width=1\linewidth]{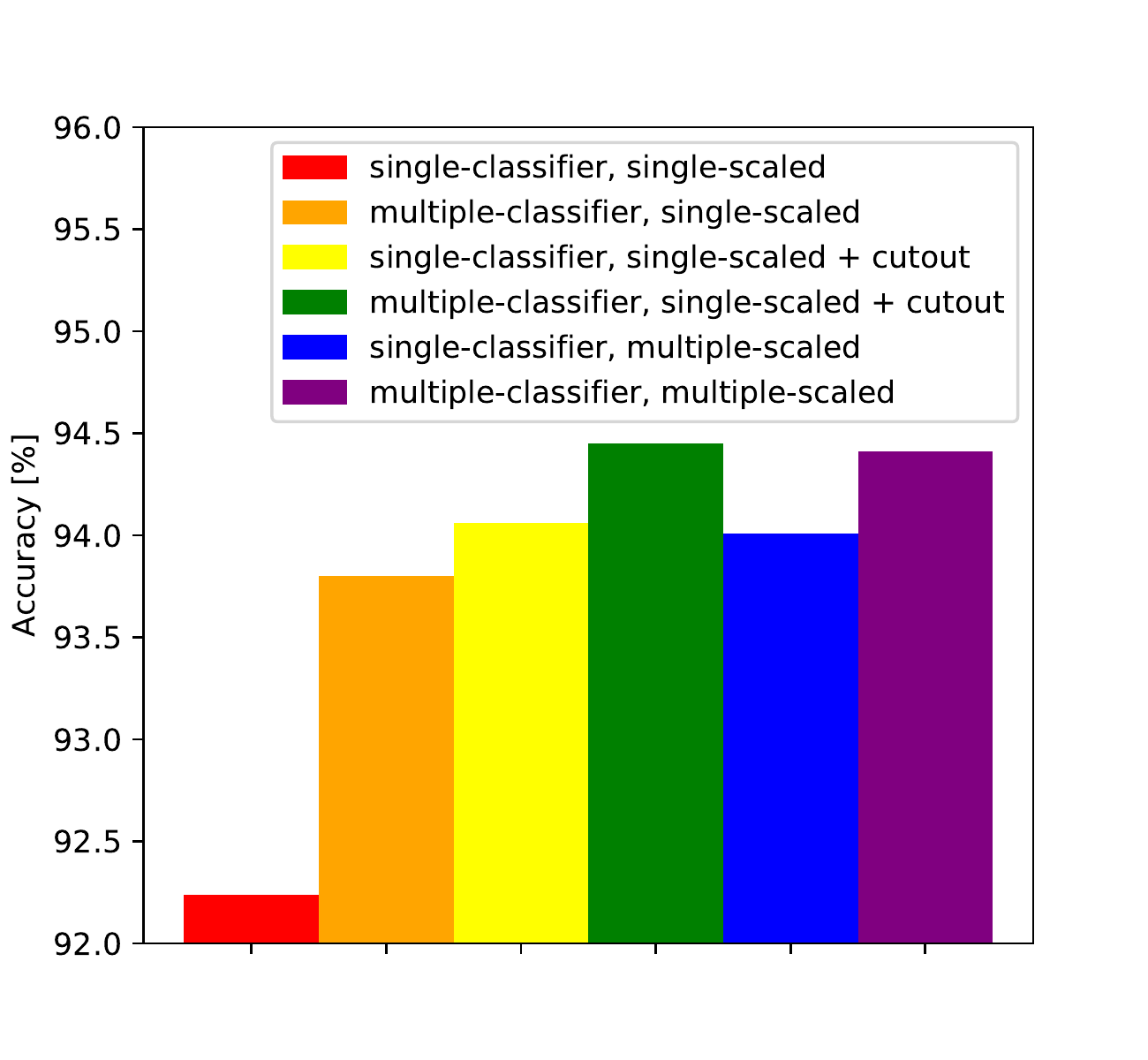}
\end{center}
   \caption{\textbf{Ablation study:} the accuracy of a single-classifier, single-scaled trained model depends on the employed architectural setting during the search phase.}
\label{fig:search}
\end{figure}

In order to asses the impact of the previously presented techniques, we have selected a 5-layer, 16-initial channels, single-classifier (\ie a singular exit point, located after the last layer), single-scaled network to represent our baseline model. Additionally, during the training phase the same network architecture was implemented, the only difference being represented by the MS-RANAS cell structure, obtained through different search processes. The results are presented in Figure \ref{fig:search}.

In terms of data preprocessing, we implemented the findings of DeVries and Taylor~\cite{cutout}, namely the use of cutout (\ie randomly masking out square regions of the input in the training phase). As shown in Figure \ref{fig:search}, applying this technique solely during the search process led to a considerable increase in accuracy when compared to the baseline network. 

Secondly, we introduced early classifiers during the searching phase, the classifiers being included on each layer except for the first one. While we could have integrated classifiers on the first layer as well, we decided not to do it, as that might have over-optimised features for the first classifier excessively early with respect to the total depth. When using only one scale during the search process, the use of early exits largely improves the overall accuracy of the network, as seen in Figure \ref{fig:search}. We postulate that this behaviour is linked with the MS-RANAS cells being forced to be more agile in order to offer good predictions in the early stages. Another aspect that we have to take into account is the relationship between adding intermediate classifiers and widening the architecture. As the number of scales increases while searching for the optimal cell structure, the positive impact of multiple classifiers is reduced. A similar behaviour is observed when analysing the impact of early exits when the search process includes the usage of cutout. We therefore conclude that early exits act as regularizers for the model, applying penalties on layer activity or layer parameters during the optimisation phase.

Lastly, we observed that, by increasing the amount of scales (\eg three scales) during the search process, the network is able to offer more accurate classifications. Specifically, we obtained an increase of almost 2\% in accuracy when using three scales during searching for the architecture instead of only one, despite the fact that only one scale was used during training. We attribute that to the fact that each cell forces their predecessors to produce more qualitative results. Therefore, this behaviour is similar to implementing a deeper network, but, instead of adding more layers and, hence, expanding the depth, we add scales, thus we increase the width. The advantage of having a rather wide than deep model lies in the increasing interaction between weights, as presented in \ref{ssc:multiscale}: during backpropagation, each operation directly influences two sets of weights, as opposed to only one set for a single-scaled model.

\subsection{Anytime inference and accuracy-latency trade-off}
\label{ssc:anytime_inference}

Our main focus is the accuracy-latency trade-off, analysed under the anytime inference setting~\cite{pmlr-v22-grubb12}. Anytime prediction implies the existence of a non-deterministic computational budget $B$ per each test sample $x$. The budget is drawn from the joint distribution $P(x,B)$. We therefore aim to minimise the expected loss
\begin{equation}
\label{eq:expected}
L(f) = \mathrm{E}{[L(f(x), B)]}_{P(x,B)},
\end{equation}
where $L(\cdot)$ represents the loss function, whilst $f(\cdot)$ is our model.

\begin{figure}[th!]
\begin{center}

\includegraphics[width=1\linewidth]{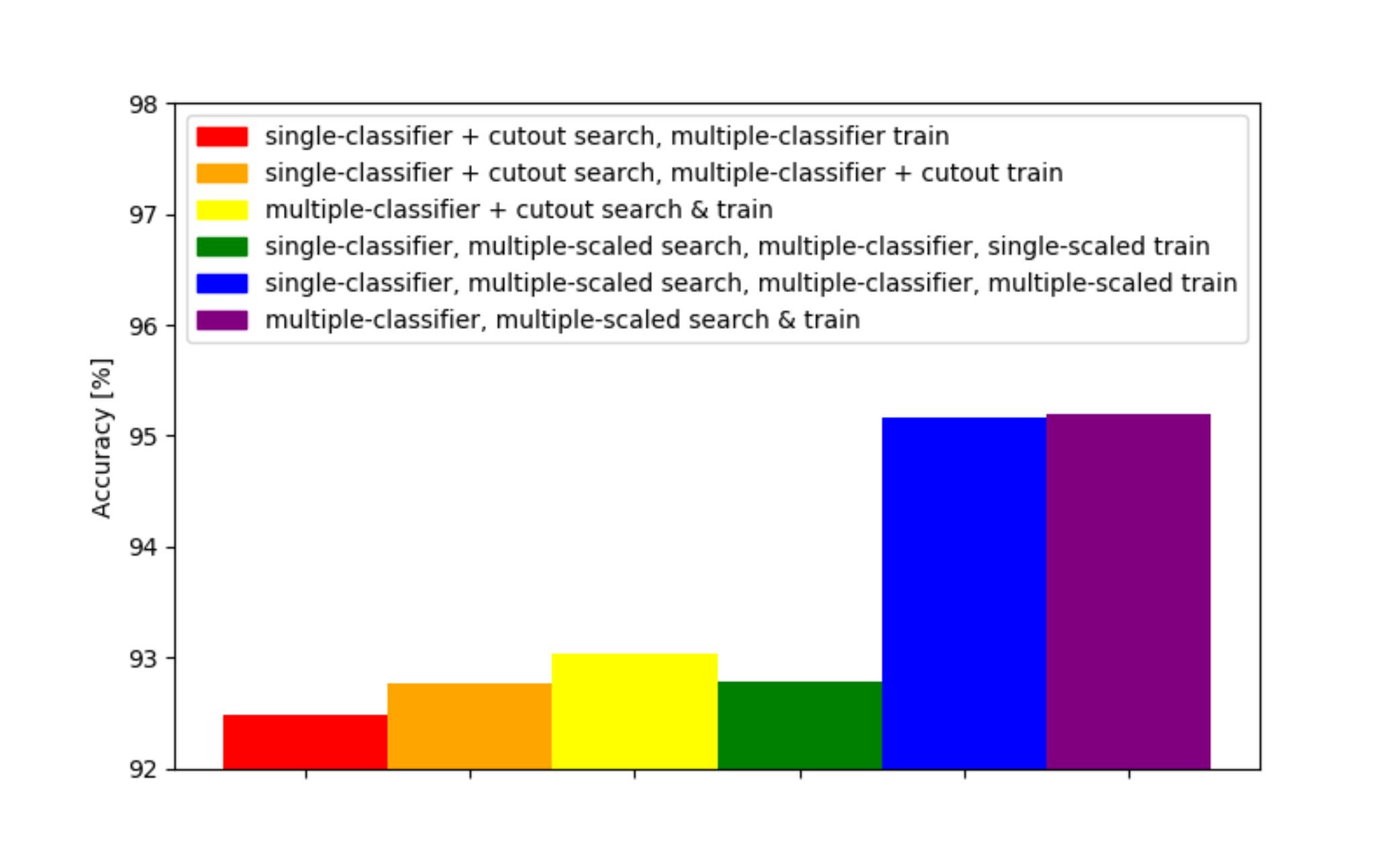}
\end{center}
\caption{\textbf{Ablation study:} the accuracy of a model depends on both the employed architectural setting during the search phase and the employed architectural setting during the train phase.}

\label{fig:train}
\end{figure}

In order to adhere to the anytime prediction principles, early classifiers had to be introduced in the final model. As mentioned in \ref{ssc:latency}, the implementation of early exits in a neural network leads to a reduction in accuracy when compared to the its single-classifier counterpart. This behaviour is visible when comparing the results depicted in Figure \ref{fig:train} and Figure \ref{fig:search}, respectively. Therefore, to alleviate this issue we exploited and further researched the findings presented in \ref{ssc:search_process}.

Firstly, due to the promising results obtained through the usage of cutout while searching for the MS-RANAS cell structure, we implemented the same technique, followed by introducing the so-obtained cell in a multi-classifier network. Whilst the cutout might have brought an improvement of the final accuracy, the insertion of early exits overshadowed it, contributing to unsatisfactory results, as shown in Figure \ref{fig:train}. We consider that this behaviour is a consequence of the network over-optimising the early features in order not be penalised by the loss function applied to the first classifiers. Hence, as the usage of cutout only during the search process was insufficient, we made additional use of it during the training phase, which led to an increase in the model's accuracy. This was followed by an additional improvement when the searching phase benefited from the implementation of multiple classifiers, thus proving the consistency of the conclusions presented in \ref{ssc:search_process}.

When we addressed the impact of multiple feature maps during the training process and both training and searching phase, respectively, a comparable behaviour, yet with a larger magnitude, was observed, as seen in Figure \ref{fig:train}. Namely, a multiple-scaled model broadly outperforms a single-scaled one when early exits are implemented. Nevertheless, driven by the encouraging results discussed in \ref{ssc:search_process}, we have introduced intermediate classifiers not only in the final architecture, but also while searching for the optimal cell, thus marginally augmenting the quality of the predictions.

\begin{figure}[th!]
\begin{center}

\includegraphics[width=1\linewidth]{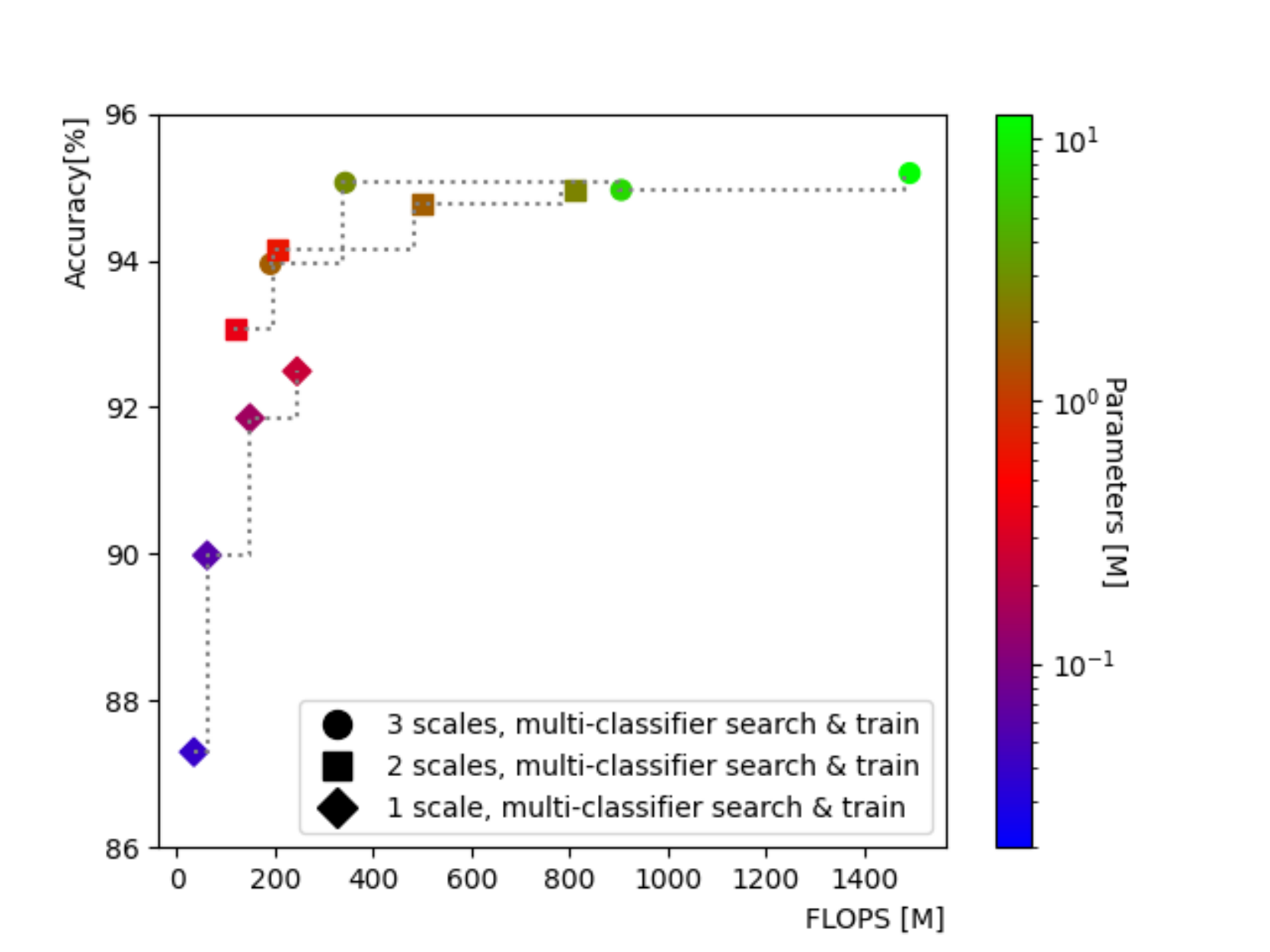}
\end{center}
      \caption{\textbf{Ablation study:} The accuracy varies in an anytime inference setup (without cutout), with respect to both floating-point operations and amount of used parameters.}
\label{fig:memory}
\end{figure}
\subsection{Memory impact in anytime inference}
\label{ssc:memory_impact}

If one application has latency as the main target and accuracy is only a second-order metric, the predictions can be taken solely from an early classifier instead of averaging out the required budget with respect to a desired accuracy. To address this issue, in Figure~\ref{fig:memory} we present a comparison of the absolute accuracy on each intermediate classifier of several architectures. We observed that a smaller difference between the errors of the first and last classifier is achieved through the usage of early classifiers and multiple scales during both searching and training phase. Namely, in the triple-scaled network there is a difference of less than 1.25\% in accuracy between the prediction obtained through the first exit and the one retrieved from the last classifier, but with the usage of only 12\% of the initial amount of millions of floating-point operations (MFLOPS) and memory requirements.

When the on-chip memory represents a third metric used to evaluate a model, it would appear that the total amount of model parameters decreases when an earlier classifier is used. Nonetheless, the storage space required to deploy the entire network is still given by the total amount of parameters, namely the value that is depicted for the last classifier. In order to de facto reduce the model parameters, one should apply the techniques presented in Section \ref{ssc:model_compression}, their implementation being outside the scope of this paper.

\begin{figure}[h]
\centering
\begin{minipage}[t]{.49\textwidth}
   \includegraphics[width=\linewidth]{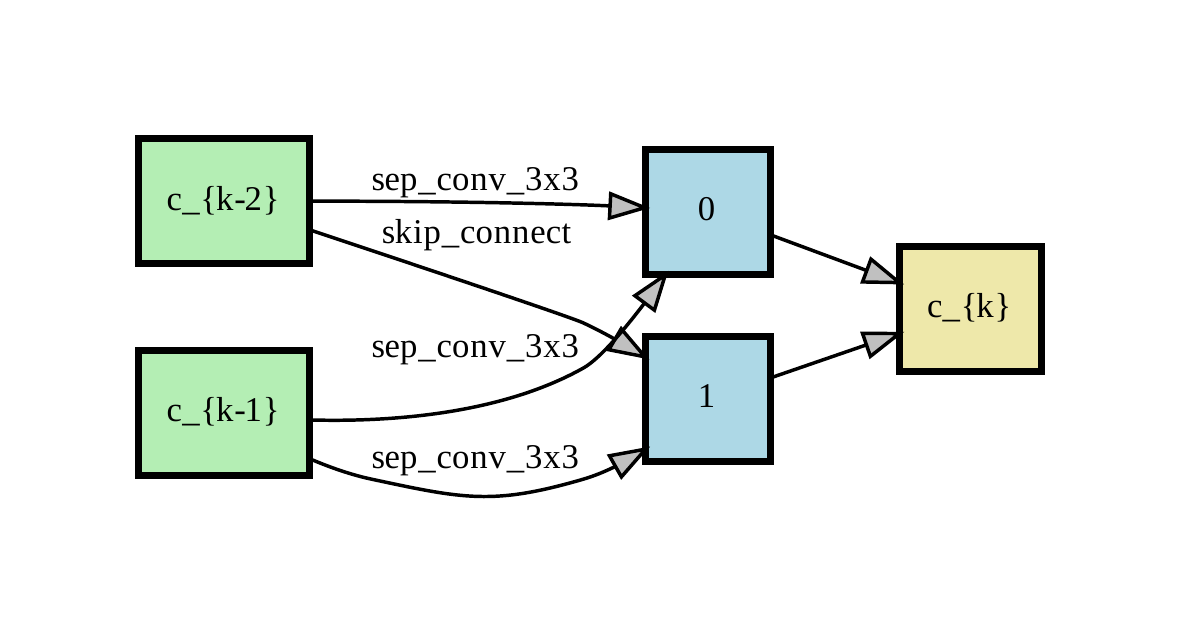}
   \caption{Example of\protect\\ 2-nodes MS-RANAS \textit{normal} cell learned on CIFAR-10.
   }
\label{fig:normal_cell}

\end{minipage}\qquad
\begin{minipage}[t]{.49\textwidth}

   \includegraphics[width=\linewidth]{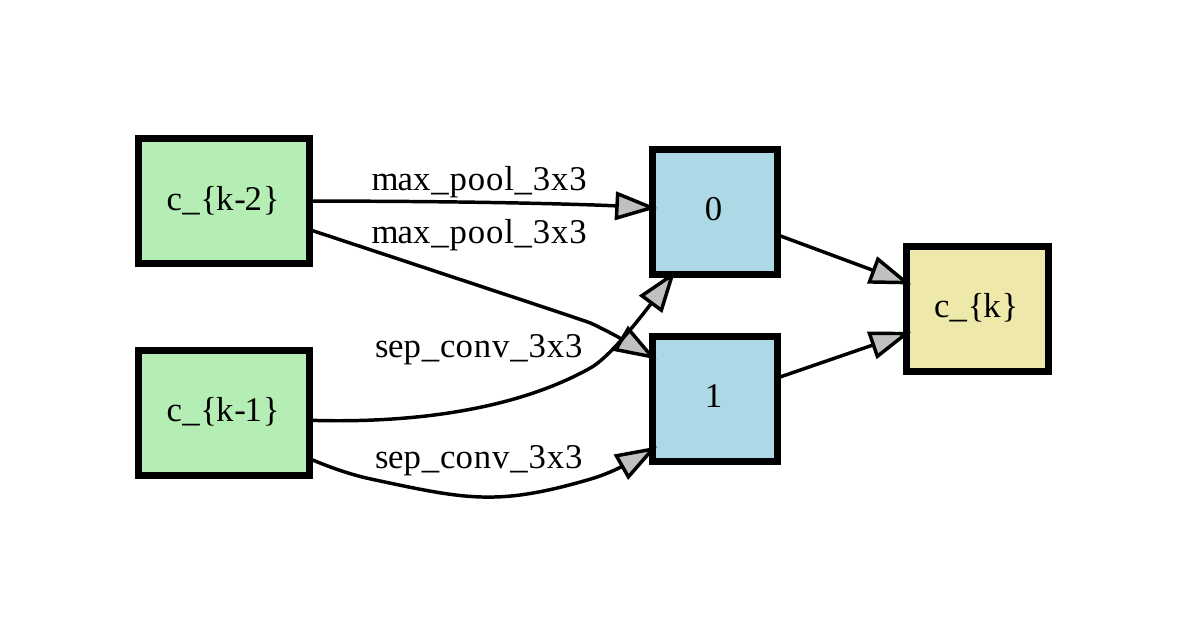}
   \caption{Example of\protect\\ 2-nodes MS-RANAS \textit{reduction} cell learned on CIFAR-10.
   }
\label{fig:reduction_cell}
\end{minipage}
\end{figure}

\subsection{Comparison with state-of-the-art}
\label{ssc:state-of-the-art}

In order to compare our MS-RANAS to the current state-of-the-art models in the field of anytime prediction, the network size (\ie number of parameters) has been chosen to match the dimension of the \textit{handcrafted} MSDNet network presented by Huang~\etal~\cite{msdnet}, thus allowing for an impartial comparison, whilst the size of the GHN model introduced by Zhang~\etal~\cite{ghn} was unavailable. Given this constraint, we could not focus on a true three-party trade-off between accuracy, computational complexity, and memory requirements, therefore the latter became a hard limitation instead of an adjustable property of our model. This led to the usage of two intermediate nodes in each cell, as opposed to the four proposed by Liu~\etal~\cite{darts}. An additional benefit of this decision was that, compared to their one-shot architecture search strategy, our search cost in terms of GPU days was further reduced. This is justified by the fact that decreasing the amount of nodes within an MS-RANAS cell reduces the number of edges in the acyclic graph, which translates into a smaller search space on which the neural architecture search algorithm operates. The final MS-RANAS cell structure is presented in Figure~\ref{fig:normal_cell}, together with the reduction cell shown in Figure~\ref{fig:reduction_cell}. For the same reason of maintaining the memory requirements within the aforementioned bounds, the final architecture is comprised of only three scales and seven layers, thus also allowing for the introduction of multiple early exits.

\begin{figure}[h]
   \includegraphics[width=\linewidth]{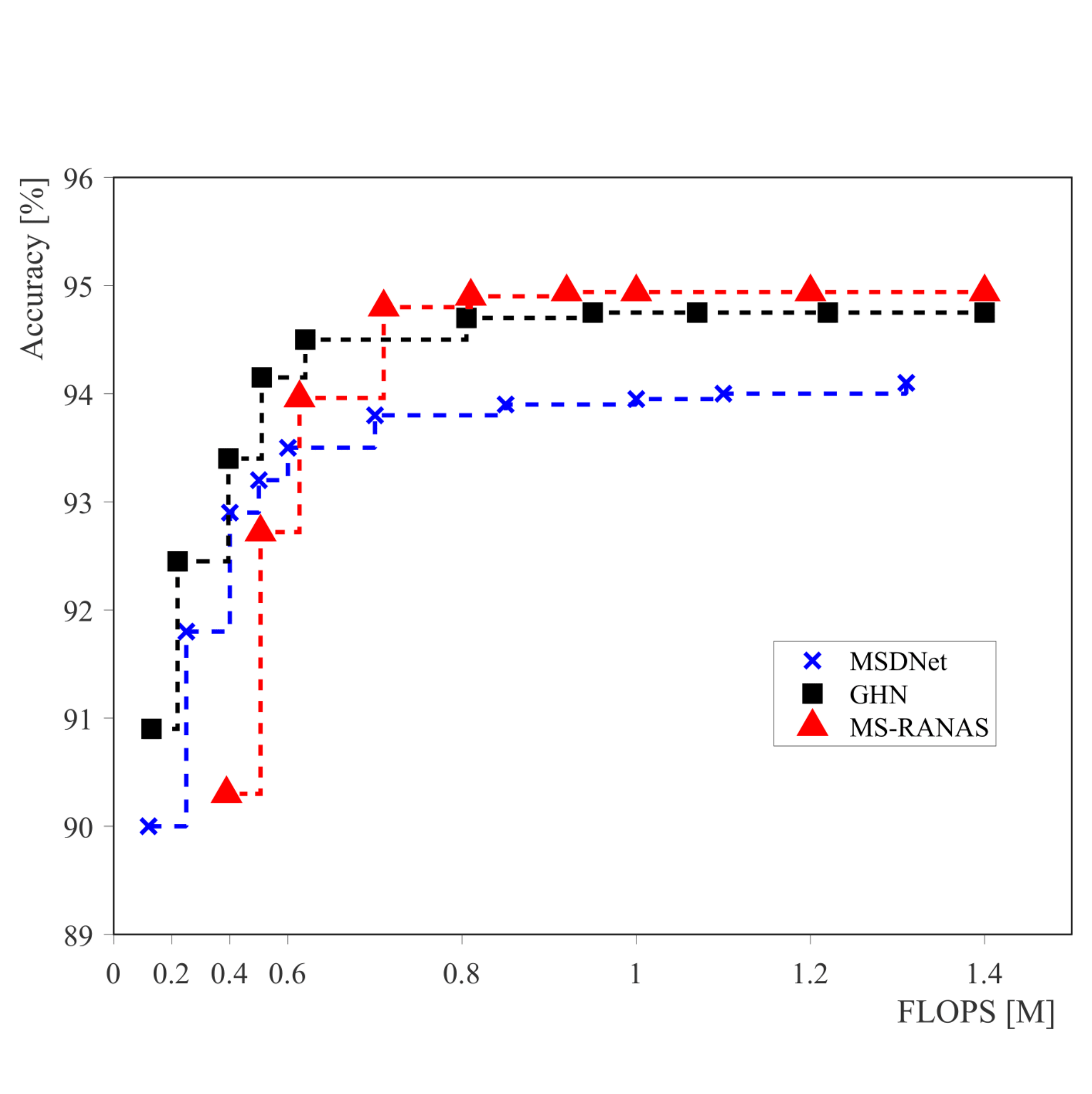}
   \caption{\textbf{CIFAR-10 anytime prediction results.} Above 0.7 MFLOPS, MS-RANAS  provides the best results. For both state-of-the-art MSDNet~\cite{msdnet} and GHN~\cite{ghn} we are presenting their best reported results out of several times more runs than for ours.
   }
\label{fig:anytime}
\end{figure}

\begin{figure}

   \includegraphics[width=\linewidth]{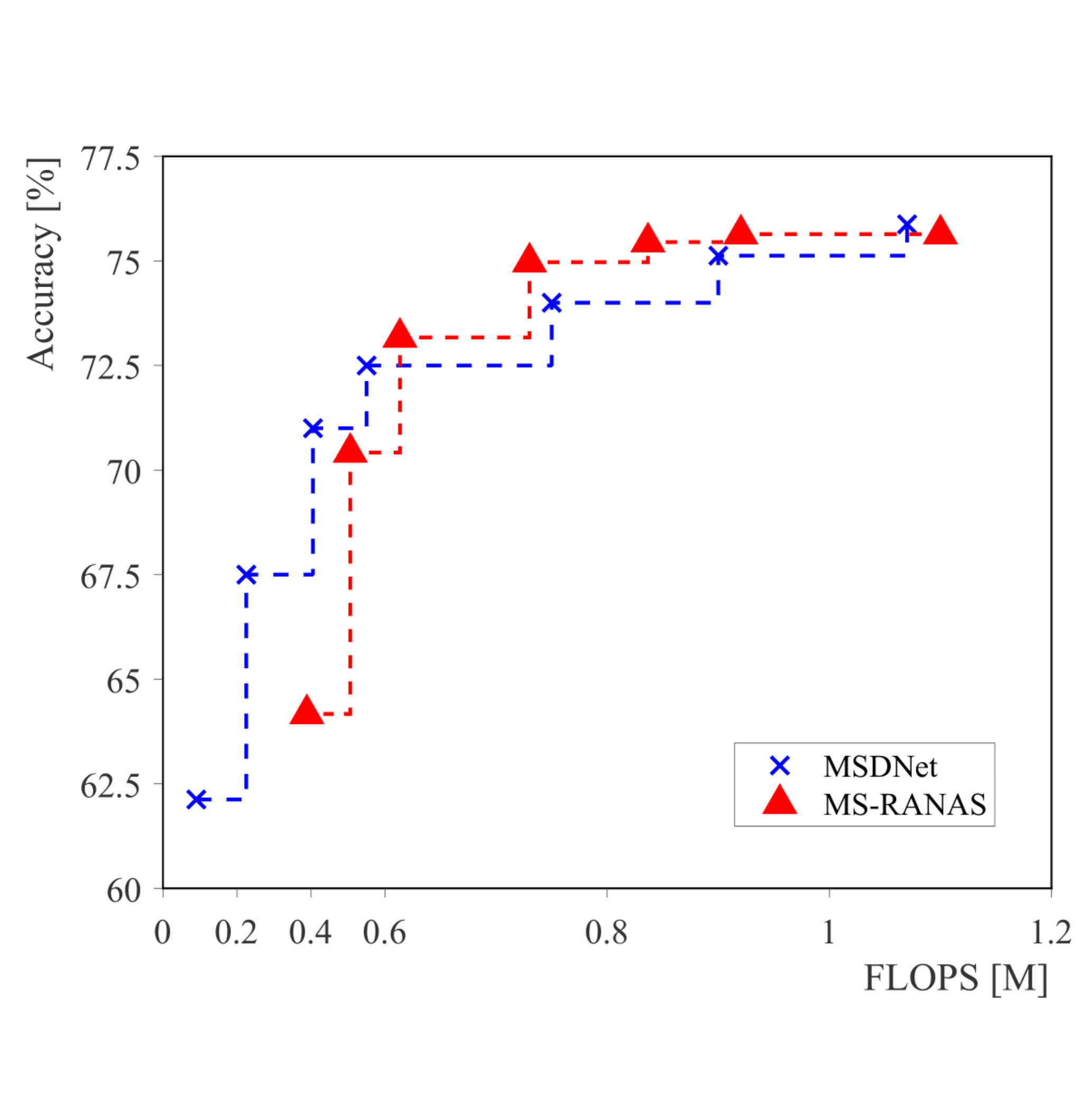}
   \caption{\textbf{CIFAR-100 anytime prediction results.} Above 0.62 MFLOPS, MS-RANAS outperforms MSDNet~\cite{msdnet}. For MSDNet we are presenting their best reported result out of several times more runs than for ours.
   }
\label{fig:anytimeCIF100}
\end{figure}

Under this specific setting, we analysed the relation between the amount of MFLOPS and the accuracy of the network on CIFAR-10, presented in Figure~\ref{fig:anytime}. As Zhang~\etal~\cite{ghn} reported the usage of cutout during GHN training, we followed their approach and used also cutout, which allowed us to successfully surpass their results as well. We also outperformed MSDNet~\cite{msdnet} after 0.7 MFLOPS. When compared to the state-of-the-art in anytime classification on CIFAR-100~\cite{cifar10}, our model outperformed MSDNet after 0.6 MFLOPS, as shown in Figure~\ref{fig:anytimeCIF100}. MS-RANAS therefore achieved scalable, competitive results in terms of accuracy-latency trade-off with handcrafted (MSDNet) or optimised (GHN) state-of-the-art architectures, while employing a limited amount of parameters.

\section{Conclusions and future work}

We introduced MS-RANAS, a novel approach towards automating the generation of efficient neural networks. Its flexibility in terms of intermediate nodes per cell, scales throughout the architecture, and positioning of early exits allows for its usage in a large variety of applications. We showed that the process of searching for the optimal MS-RANAS cell can be augmented through the usage of cutout, multiple scales, and intermediate classifiers, thus obtaining a more accurate network despite the absence of these techniques in the final model. When the main aim of the system is the trade-off between speed and accuracy (\eg{anytime prediction}), MS-RANAS network outperforms the existing state-of-the-art architectures, either handcrafted or not, while requiring a comparable amount of parameters. This makes our work an ideal candidate for object recognition tasks on an edge computing setting, where there exists a joint focus on accuracy, latency, and on-chip memory.

We plan to further analyse two directions which could lead to improved results. First of all, as we have presented in subsection~\ref{ssc:model_compression}, several techniques such as pruning and quantisation could be used to further reduce the dimensionality of the network. Secondly, in order to allow for accurate predictions under a scarce computational budget, utilizing the area under accuracy-MFLOPS curve as a selection criteria during the architecture search phase might be beneficial, thus generating a model efficiently operating under any computational constraints. Additionally, directly optimising latency instead of using a surrogate (~\ie{FLOPS}), as proposed by Tan~\etal~\cite{mnasnet}, could lead to a better accuracy-latency trade-off.

{\small
\bibliographystyle{ieee_fullname}
\bibliography{egbib}
}

\end{document}